\documentclass{article}
\usepackage{spconf}
\usepackage{amsmath}
\usepackage{graphicx}
\usepackage{amsthm}
\usepackage{amssymb}
\usepackage{color}
\usepackage{subcaption}
\usepackage{array}
\usepackage{enumitem}
\usepackage{booktabs}
\usepackage{multirow} 
\usepackage{caption}
\usepackage{subcaption}
\usepackage{makecell}

\title{Real-time 2D/3D Registration via CNN Regression}

\name{Shun Miao$^{\dagger \star}$ \qquad Z. Jane Wang$^{\dagger}$ \qquad Yefeng Zheng$^{\star}$ \qquad Rui Liao$^{\star}$}
\address{
        $^{\dagger}$Electrical and Computer Engineering, University of British Columbia, Canada \\
        $^{\star}$ Medical Imaging Technologies, Siemens Healthcare, USA \\  }

\begin{document}

\maketitle

\begin{abstract}
In this paper, we present a Convolutional Neural Network (CNN) regression approach for real-time 2-D/3-D registration. Different from optimization-based methods, which iteratively optimize the transformation parameters over a scalar-valued metric function representing the quality of the registration, the proposed method exploits the information embedded in the appearances of the Digitally Reconstructed Radiograph and X-ray images, and employs CNN regressors to directly estimate the transformation parameters. The CNN regressors are trained for local zones and applied in a hierarchical manner to break down the complex regression task into simpler sub-tasks that can be learned separately. Our experiment results demonstrate the advantage of the proposed method in computational efficiency with negligible degradation of registration accuracy compared to intensity-based methods. 
\end{abstract}


\def\refEqn#1{Eqn.~(\ref{#1})}
\def\refFig#1{Fig.~\ref{#1}}
\def\refTable#1{Table~\ref{#1}}

\section{Introduction}
\label{sec:intro}

2-D/3-D registration represents one of the key enabling technologies in medical imaging and image-guided interventions~\cite{liao2013review}. 
It can bring the pre-operative 3-D data and intra-operative 2-D data into the same coordinate system, to facilitate accurate diagnosis and/or provide advanced image guidance.
The pre-operative 3-D data generally includes Computed Tomography (CT), Cone-beam CT (CBCT), Magnetic Resonance Imaging (MRI) and Computer Aided Design (CAD) model of medical devices, while the intra-operative 2-D data is dominantly X-ray images.
In this paper,  we focus on registering a 3-D X-ray attenuation map provided by CT or CBCT with a 2-D X-ray image in real-time.

Although 2-D/3-D registration is a widely adopted technology in medical imaging, real-time 2-D/3-D registration with sub-millimeter accuracy remains a great challenge.
Most existing 2-D/3-D registration methods in the literature are optimization-based, in which the transformation parameters are iteratively updated to optimize an objective function reflecting the quality of the registration.
Depending on the objective function to be optimized, optimization-based methods can be further divided into intensity-based and feature-based methods \cite{markelj2012review}.
In intensity-based methods, a simulated X-ray image, referred to as Digitally Reconstructed Radiograph (DRR), is derived from the 3-D X-ray attenuation map by simulating the attenuation of virtual X-rays~\cite{liao2013automatic}\cite{miao2013toward}.
An optimizer is employed to maximize an intensity-based similarity measure between the DRR and X-ray images.
Intensity-based methods are widely adopted mainly because of their high accuracy \cite{wu2009evaluation}. 
However, they often involve a large number of evaluations of the similarity measure, each requiring a high computational cost in rendering the DRR, and as a result are typically not suitable for real-time applications.
In comparison, feature-based methods calculate similarity measures efficiently from geometric features extracted from the images, e.g., corners, lines and segmentations~\cite{groher2007segmentation}\cite{duong2009curve}, and therefore have a higher computational efficiency than intensity-based methods.
One potential drawback of feature-based methods lies in the fact that they heavily rely on accurate detection of geometric features, which by itself could be a challenging task.
Errors from the feature detection step are inevitably propagated into the registration result~\cite{mahfouz2005effect}, making feature-based methods in general less accurate~\cite{mclaughlin2002comparison}.

In this paper, a Convolutional Neural Network (CNN) regression approach is presented for real-time 2-D/3-D registration.
The effectiveness of CNN has been shown in a wide range of computer vision tasks~\cite{krizhevsky2012imagenet}, but to the best of the authors' knowledge, it has not been reported for  2-D/3-D registration.
We rely on the strong non-linear modeling capability of CNN to directly estimate the transformation parameters from the appearance of DRR and X-ray images.
Comparing to intensity-based methods, which maps the images to a scalar-valued metric function, the proposed method better exploits the information embedded in the images for more efficient parameter update.
Therefore, accurate 2-D/3-D registration can be achieved with very few DRR renderings, making the proposed method highly computationally efficient and suitable for real-time applications.

\section{Problem Formulation}
\label{sec:dnn2d3d}


\subsection{3-D Transformation Parameterization}

\begin{figure}
	\centering
	\includegraphics[width=\linewidth]{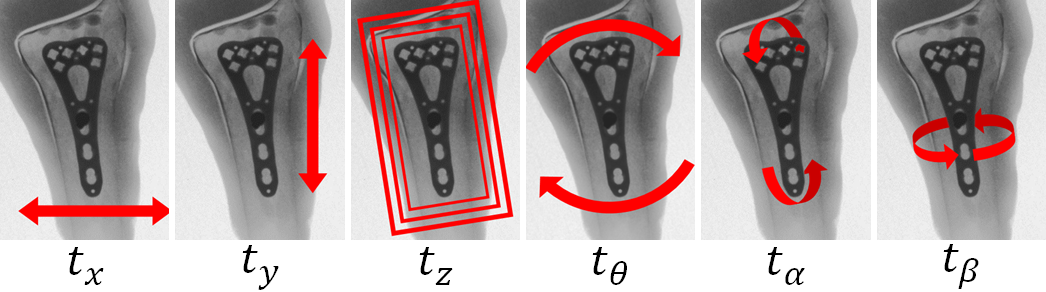}
	\caption{Effects of the 6 transformation parameters}
	\label{fig:paramseffects}
\end{figure}

A rigid-body 3-D transformation $T$ can be parameterized by a vector $\boldsymbol{t}$ with 6 components.
In our approach, we parameterize the transformation by 3 in-plane and 3 out-of-plane transformation parameters~\cite{kaiser20142d}, as shown in Fig. \ref{fig:paramseffects}.
In particular, in-plane transformation parameters include 2 translation parameters, $t_x$ and $t_y$, and 1 rotation parameter, $t_\theta$.
The effects of in-plane transformation parameters are approximately 2-D rigid-body transformations. 
Out-of-plane transformation parameters include 1 out-of-plane translation parameter, $t_z$, and 2 out-of-plane rotation parameters, $t_\alpha$ and $t_\beta$.
The effects of out-of-plane translation and rotations are scaling and shape changes, respectively. 

\subsection{2-D/3-D Registration via Regression}

We denote the X-ray image with transformation parameters $\boldsymbol{t}$ as $I_{\boldsymbol{t}}$.
The inputs for 2-D/3-D registration are: 1) a 3-D object described by its X-ray attenuation map $J$, 2) an X-ray image $I_{\boldsymbol{t}_{gt}}$, where $\boldsymbol{t}_{gt}$ denotes the unknown ground truth transformation parameters, and 3) initial transformation parameters $\boldsymbol{t}_{ini}$.
The goal of 2-D/3-D registration is to estimate $\boldsymbol{t}_{gt}$ from the inputs.
It can be formulated as a regression problem, where a set of regressors $\boldsymbol{f}(\cdot)$ are trained to reveal the mapping from a feature $X(\boldsymbol{t}_{ini}, I_{\boldsymbol{t}_{gt}})$ extracted from the inputs to the difference between $\boldsymbol{t}_{ini}$ and $\boldsymbol{t}_{gt}$, as long as it is within a pre-defined range $\boldsymbol{\epsilon}$:
\begin{equation}
	\boldsymbol{t}_{gt} - \boldsymbol{t}_{ini} \approx \boldsymbol{f}\big(X(\boldsymbol{t}_{ini}, I_{\boldsymbol{t}_{gt}}) \big),\quad
	\forall \boldsymbol{t}_{gt} - \boldsymbol{t}_{ini} \in \boldsymbol{\epsilon}.
	\label{eqn:regexe}
\end{equation}
An estimation of $\boldsymbol{t}_{gt}$ is then obtained by applying the regressors and incorporating the result into $\boldsymbol{t}_{ini}$:
\begin{equation}
	\hat{\boldsymbol{t}}_{gt} = \boldsymbol{t}_{ini} + \boldsymbol{f}\big(X(\boldsymbol{t}_{ini}, I_{\boldsymbol{t}_{gt}}) \big).
\end{equation}
It is worth noting that the range $\boldsymbol{\epsilon}$ in \refEqn{eqn:regexe} is equivalent to the capture range of optimization-based registration methods.
Based on \refEqn{eqn:regexe}, our problem formulation can be expressed as designing a feature extractor $X(\cdot)$ and training regressors $\boldsymbol{f}(\cdot)$, such that
\begin{equation}
	\delta \boldsymbol{t} \approx \boldsymbol{f}\big(X(\boldsymbol{t}, I_{\boldsymbol{t} + \delta \boldsymbol{t}}) \big), \quad \forall \delta \boldsymbol{t} \in \boldsymbol{\epsilon}.
	\label{eqn:regform}
\end{equation}
In the next section, we will discuss in details 1) how the feature $X(\boldsymbol{t}, I_{\boldsymbol{t} + \delta \boldsymbol{t}})$ is calculated and 2) how the regressors $\boldsymbol{f}(\cdot)$ are designed, trained and applied.

\section{Method}

\subsection{Feature Extraction}

\begin{figure}
        \centering
        \includegraphics[width=0.9\linewidth]{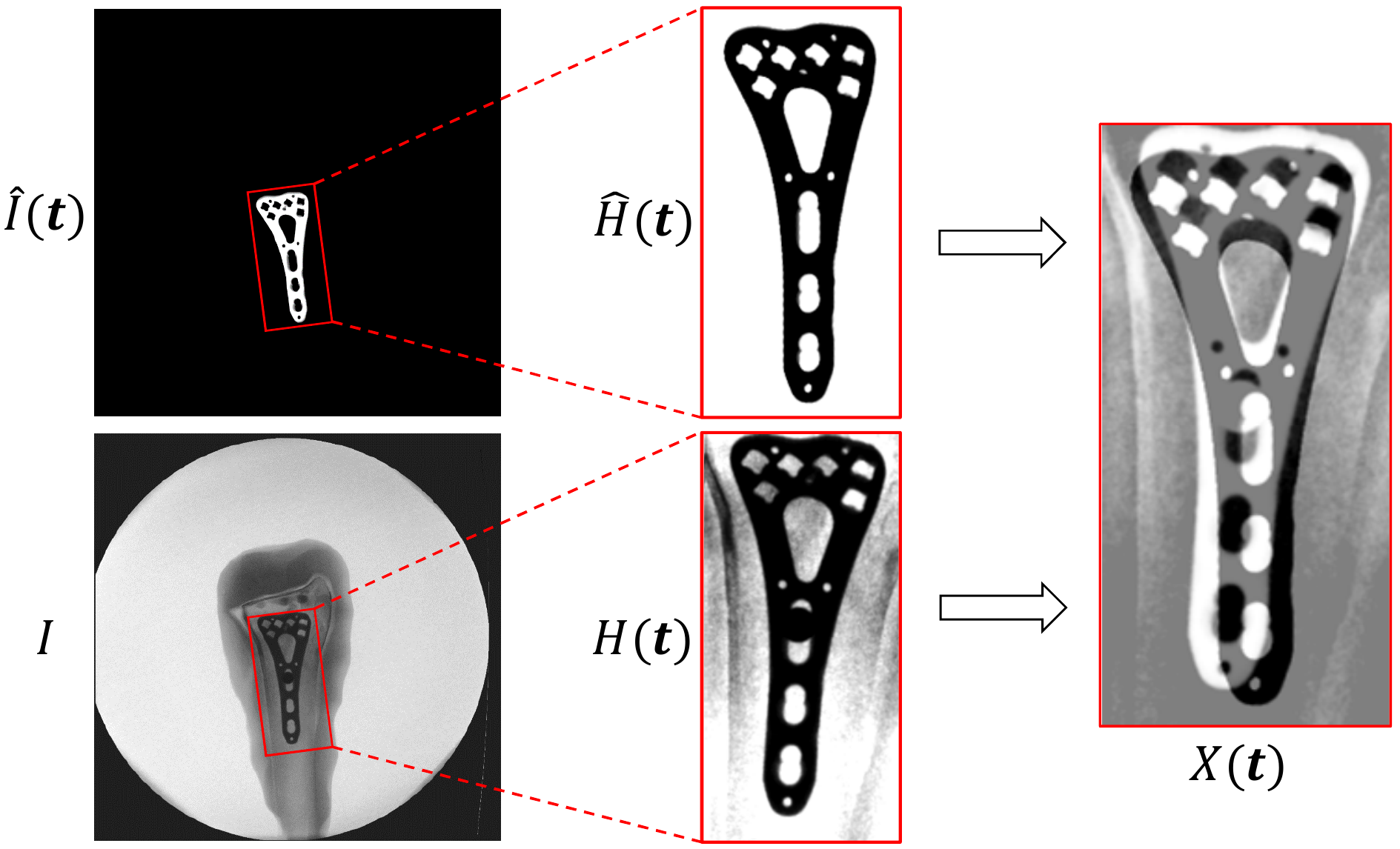}
        \caption{Feature extraction from the DRR and X-ray images}
        \label{fig:feature}
\end{figure}

We compute the residual between the DRR with transformation parameters $\boldsymbol{t}$, denoted by $I_{\boldsymbol{t}}$, and the X-ray image $I_{\boldsymbol{t}+ \delta \boldsymbol{t}}$, and use it as the feature for regression.
The residual is computed within an ROI around the target object in the DRR, determined by $\boldsymbol{t}$, as shown in \refFig{fig:feature}.
A ROI can be described by $(\boldsymbol{q}, w, h, \phi)$, denoting the ROI's center, width, height and orientation, respectively.
The center $\boldsymbol{q}$ is the 2-D projection of gravity center of the target object using transformation parameters $\boldsymbol{t}$.
The width and height are calculated as $w = w_0 \cdot D /t_z $ and $h = h_0 \cdot D /t_z $, respectively, where $w_0$ and $h_0$ are the size of the ROI in mm and $D$ is the distance between the X-ray source and detector.
The orientation $\phi = t_\theta$, so that it is always aligned with the object.
We define an operator $H^{\boldsymbol{t}} (\cdot)$ that extracts the image patch in the ROI determined by $\boldsymbol{t}$ , and re-sample it to a fixed size (156$\times$300 in our experiment).
The feature used for  regression is then calculated as
\begin{equation}
    X(\boldsymbol{t}, I_{\boldsymbol{t} + \delta \boldsymbol{t}}) = H^{\boldsymbol{t}} (I_{\boldsymbol{t}}) - H^{\boldsymbol{t}} (I_{\boldsymbol{t} + \delta \boldsymbol{t}}).
    \label{eqn:feature}
\end{equation}

\subsection{Hierarchical Regression}
\label{subsubsec:hierreg}
\begin{figure}
        \centering
        \includegraphics[width=\linewidth]{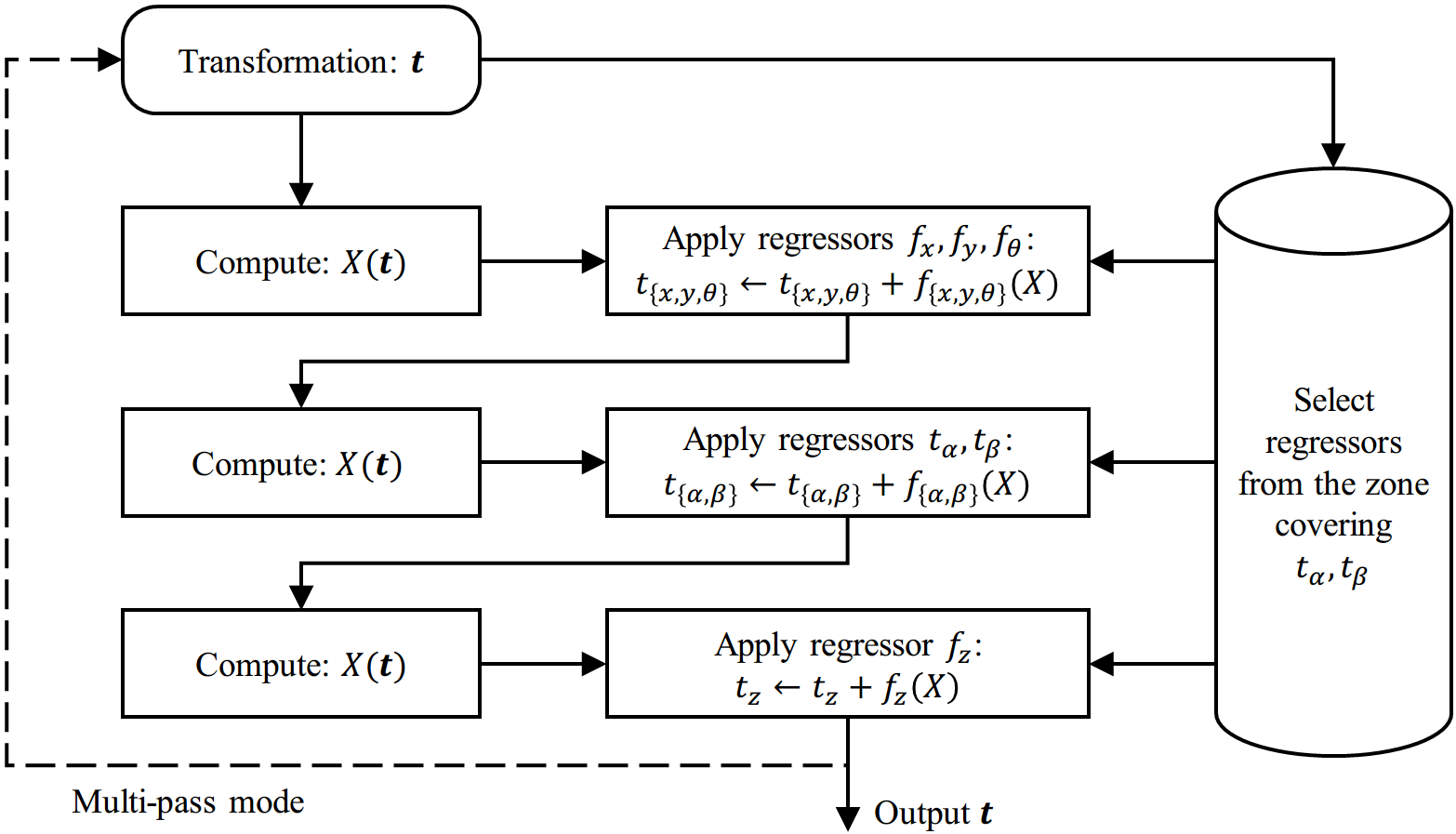}
        \caption{Workflow of the hierarchical regression strategy}
        \label{fig:workflow}
\end{figure}

Our goal is to train 6 regressors $\boldsymbol{f} = \{f_x,f_y,f_z, f_\theta, f_\alpha, f_\beta\}$ to reveal the correlation between $X$ and $\delta \boldsymbol{t}$.
Considering that $X$ only contains 2-D information, the mapping from $X$ to $\delta \boldsymbol{t}$ could be very complex.
To reduce the complexity of the regression problems, we carry out the following hierarchical regression steps.
The steps are also illustrated in the workflow diagram shown in Fig.~\ref{fig:workflow}. 

We first partition the parameter space spanned by $t_\alpha$ and $t_\beta$ with a 18$\times$18 grid, each covering a 20$^\circ$$\times$20$^\circ$ zone.
Six regressors are trained  for each individual zone to solve 2-D/3-D registration problems with initial $t_\alpha$ and $t_\beta$ in this zone.
2-D/3-D registration tasks are dispatched into corresponding zones, according to their initial values of $t_\alpha$ and $t_\beta$.
Using this strategy, each regressor only needs to reveal the correlation between $X$ and $\delta \boldsymbol{t}$ for a small range of $t_\alpha$ and $t_\beta$ (i.e., 20$^\circ$), making the regression problems much simpler.

We then divide the 6 regressors into 3 groups, $\{f_x,f_y,f_\theta\}$, $\{f_\alpha,f_\beta \}$ and $\{f_z\}$, and regress them hierarchically.
Among the 3 groups, the parameters in Group 1 are considered to be the easiest to be estimated, because they cause simple and dominant rigid-body 2-D transformation of the object in the projection image and are less affected by the variations of other parameters.
The parameter in Group 3 is the most difficult one to be estimated, because it only causes subtle scaling of the object in the projection image.
The difficulty in estimating parameters in Group 2 falls in-between.
Therefore, we regress the 3 groups of parameters sequentially, from the easiest group to the most difficult one.
After a group of parameters are regressed, the feature $X(\boldsymbol{t}, I_{\boldsymbol{t} + \delta \boldsymbol{t}})$ is re-calculated using the already-estimated parameters for the regression of the parameters in the next group.
This way the regression for the current group becomes less complicated by removing the compounding factors coming from those parameters in the previous groups.

The above hierarchical regressors can be applied once (\textit{single-pass} mode) or multiple times (\textit{multi-pass} mode).
The multi-pass mode repeats the regression process for multiple iterations, with the result of the previous iteration being used as the starting position for the current iteration.

\subsection{Convolutional Neural Network for Regression}

\subsubsection{Network Structure}

\begin{figure}
	\centering
	\includegraphics[width=\linewidth]{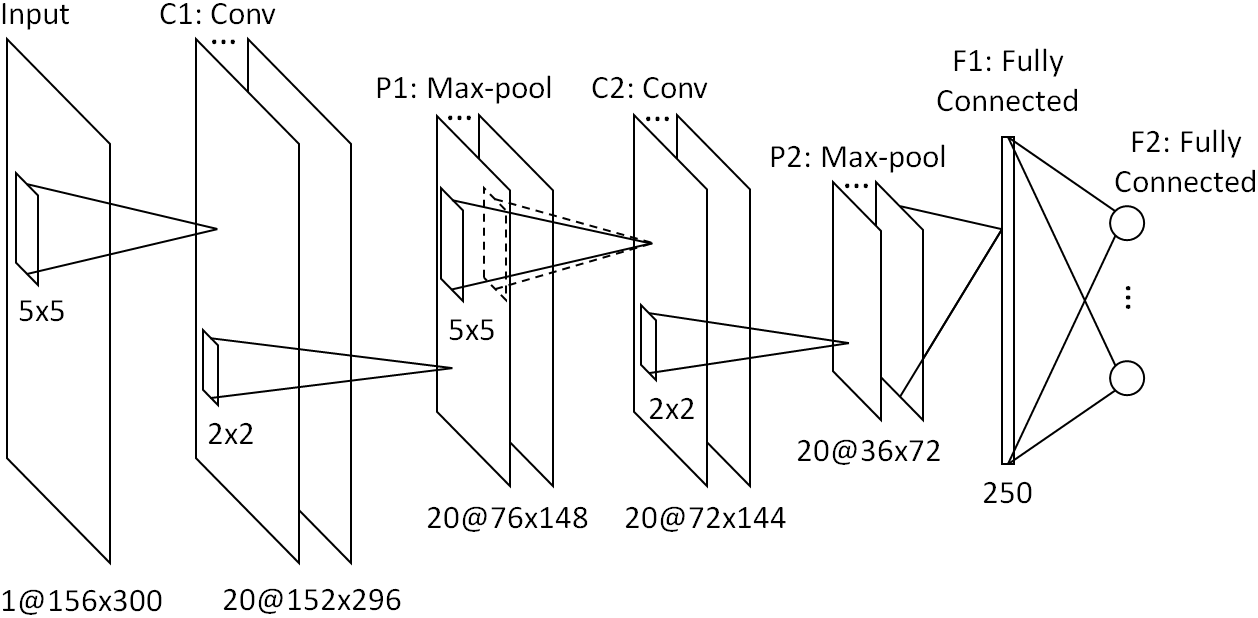}
	\caption{Structure of the multi-task learning convolutional neural network.}
	\label{fig:cnn}
\end{figure}

One CNN regression model with the architecture shown in \refFig{fig:cnn} is trained for each group in each zone.
The input of the CNN regression model is a 156$\times$300 image, computed following \refEqn{eqn:feature}.
The CNN consists of five layers, including two 5$\times$5 convolutional layers (C1 and C2), each followed by a $2\times2$ max-pooling layers (P1 and P2) with a stride of 2, and a fully-connected layer (F1) with 250 Rectified Linear Unit (ReLU) activations neurons. 
The output layer (F2) is fully-connected to F1, with each output node corresponding to one parameter in the group.

\subsubsection{Training}

The CNN regression models are trained exclusively on synthetic X-ray images, because they provide reliable ground truth labels with little needs on laborious manual annotation, and the number of real X-ray images could be limited.
For each group in each zone, we randomly generate 25,000 pairs of $\boldsymbol{t}$ and $\delta \boldsymbol{t}$.
The parameters $\boldsymbol{t}$ follow a uniform distribution with $t_\alpha$ and $t_\beta$ constrained in the zone. 
The parameter errors $\delta \boldsymbol{t}$ for Group 1 follow a zero mean uniform distribution over ranges of 3.0 mm, 3.0 mm, 30.0 mm, 6$^\circ$, 30$^\circ$ and 30$^\circ$. 
The unform distribution ranges of $\delta t_x$, $\delta t_y$ and $\delta t_\theta$ are reduced for Group 2 to 0.4 mm, 0.4 mm and 1.0$^\circ$, because they are close to zero after the regressors in the Group 1 are applied.
For the same reason, the distribution ranges of $\delta t_\alpha$ and $t_\beta$ are reduced for Group 3 to 1.5$^\circ$ and 1.5$^\circ$.
For each pair of $\boldsymbol{t}$ and $\delta \boldsymbol{t}$, a synthetic X-ray image $I_{\boldsymbol{t} + \delta \boldsymbol{t}}$ is generated and the feature $X(\boldsymbol{t}, I_{\boldsymbol{t} + \delta \boldsymbol{t}})$ is calculated following \refEqn{eqn:feature}.

The objective function to be minimized during the training is defined as:
\begin{equation}
\Phi = \frac{1}{K} \sum_{i=1}^K { \|\boldsymbol{y}_i - \boldsymbol{f}(X_i; \mathbf{W}) \|_2^2},
\label{eqn:global}
\end{equation}
where $K$ is the number of training samples, $\boldsymbol{y}_i$ is the label for the $i$-th training sample, $\mathbf{W}$ is a vector of weights to be learned, $\boldsymbol{f}(X_i; \mathbf{W})$ is the output of the regression model parameterized by $\mathbf{W}$ on the $i$-th training sample.
The weights $\mathbf{W}$ are learned using Stochastic Gradient Descent (SGD)~\cite{krizhevsky2012imagenet}, with a batch size of 64, momentum of $m=0.9$ and weight decay of $d=0.0001$ .
The learning rate $\kappa_i$ is decayed in each iteration following
$\kappa_i = 0.0025 \cdot (1+0.0001 \cdot i)^{-0.75}$.
The weights are initialized using the Xavier method~\cite{glorot2010understanding}, and mini-batch SGD is performed for 32 epochs.


\section{Experiments and Results}

\subsection{Experiment Setup}

We conducted experiments on a dataset from a potential application, Virtual Implant Planning System (VIPS), which is an intraoperative application to facilitate the planning of implant placement in terms of orientation, angulation and length of the screws~\cite{vetter2014validation}.
In VIPS, 2-D/3-D registration can be performed to match the 3-D virtual implant with the fluoroscopic image of the real implant.
The dataset consists of a CAD model  of a volar plate and 7 X-ray images of the volar plate implant mounted onto a phantom model of the distal radius.
The size of the X-ray images is 1024$\times$1024 with a pixel spacing of 0.223 mm.
The 3-D CAD model was converted to a binary volume using marching cube algorithm for registration.
Ground truth transformation parameters used for quantifying registration error were generated by first manually registering the target object and then applying an intensity-based 2-D/3-D registration method using Powell's method combined with Gradient Correlation (GC).
For each X-ray image, 140 perturbations of the ground truth were generated as starting positions for 2-D/3-D registration.
The perturbation followed zero mean Gaussian distribution with standard deviations of 1.0 mm, 1.0 mm, 10.0 mm, 2$^\circ$, 10$^\circ$ and 10$^\circ$.

We compared the proposed method in \textit{three-pass} mode with three state-of-the-art intensity-based 2-D/3-D registration methods.
Powell's method was adopted as the optimizer for all evaluated intensity-based methods as its advantage in 2-D/3-D registration over other popular optimization methods has been shown in~\cite{kaiser1232014comparison}.
We evaluated two popular similarity measures, Mutual Information (MI) and GC, which have also been reported to be effective in recent literature~\cite{gendrin2012monitoring}\cite{schmid2014segmentation}.
We also merged the two methods using MI and GC to form an improved intensity-based 2-D/3-D registration method for comparison.
The combined method, referred to as MI+GC, first applies MI to bring the registration into the capture range of GC, and then applies GC to refine the registration.

The experiments were conducted on a workstation with Intel Core i7-4790k CPU, 16GB RAM and Nvidia GeForce GTX 980 GPU. For intensity-based methods, the most computationally intensive component, DRR renderer, was implemented using the Ray Casting algorithm with GPU acceleration. Similarity measures were implemented in C++ and executed in a single CPU core. Both DRR and similarity measure were only calculated within a 512$\times$512 ROI surrounding the target object, for better computational efficiency. For the proposed method, the neural network was implemented with GPU acceleration using an open-source deep learning framework, Caffe~\cite{jia2014caffe}.

\subsection{Results}

\begin{table}[t]
	\setlength{\tabcolsep}{8pt}
	\footnotesize
	\centering
	\caption{Quantitative experiment results
		including: 1) success rate, 2) mean mTREproj calculated among successful registration, and 3) average and standard deviation of running time per registration.}
	\label{tab:results}
	\begin{tabular}{cccccc}
		\toprule
		Method & Success Rate & Mean mTREproj & Running Time \\ \midrule
		MI       & 75.1\%  & 0.315 mm & 1.66$\pm$0.60 s\\
		GC       & 78.7\%  & 0.285 mm & 3.91$\pm$1.55 s\\
		MI+GC    & 92.7\%  & 0.260 mm & 4.71$\pm$1.59 s\\
		Proposed & 92.3\%  & 0.282 mm & 0.08$\pm$0.00 s\\ \bottomrule
	\end{tabular}        
\end{table}%

The registration accuracy was accessed with the mean Target Registration Error in the projection direction (mTREproj)~\cite{de2005standardized}, calculated at the 8 corners of the bounding box of the target object.
We regard mTREproj less than 1\% of the size of the target object (i.e. diagonal of the bounding box) as a successful registration, which is equivalent to 0.61 mm.
For each evaluated method, we report its success rate, mean of mTREproj of successful registrations and running time per registration.

Table~\ref{tab:results} summarizes the experiment results.
Both MI and GC resulted in relatively low success rates (75.1\% and 78.7\%), because of the low accuracy of MI and the small capture range of GC. 
By combing the advantages of MI and GC, MI+GC, achieved much higher success rate (92.7\%) and very low mTREProj (0.260 mm), suggesting that it achieves both high robustness and accuracy.
In comparison, the proposed method achieved comparable success rate (92.3\%) with a slightly higher but still similar mTREproj (0.282 mm), compared to MI+GC.
Considering that the ground truth parameters were generated using GC, which could bear a slight bias toward intensity-based methods using GC as the similarity measure, the small differences in success rate and mTREproj between the proposed method and MI+GC indicate that they achieved comparable robustness and accuracy.

In terms of speed, the 3 intensity-based methods, MI, GC and MI+GC, are in general not fast enough for real-time registration.
The fastest one, MI, took in average 1.66 s to accomplish 2-D/3-D registration, while the most accurate one, MI+GC, took in average 4.71 s.
In comparison, the proposed method achieved a significantly higher speed (0.08 s), demonstrating its significant advantage in computational efficiency.
In addition, the running time for intensity-based methods has relatively large standard deviations because the number of iterations involved in the optimization can vary for each registration depending on the starting position.
In comparison, the standard deviation of the computation time for the proposed method is almost zero, showing that it can provide a real-time registration with a constant frame rate.

\section{Conclusion}

In this paper, we presented a real-time 2-D/3-D registration approach based on CNN regression.
We showed that 2-D/3-D registration can be efficiently solved by training CNN regressors to reveal the mapping from image residual to transformation parameter residual.
We also validated via experiments that the proposed method achieved significantly higher computational efficiency than intensity-based methods, with negligible degradation of registration accuracy.

\bibliographystyle{IEEEbib}
\bibliography{refs}

\end{document}